\documentclass[10pt,twocolumn,letterpaper]{article}

\usepackage[pagenumbers]{wacv} 

\usepackage{pifont}
\usepackage{makecell}

\definecolor{wacvblue}{rgb}{0.21,0.49,0.74}
\usepackage[pagebackref,breaklinks,colorlinks,allcolors=wacvblue]{hyperref}

\def\wacvPaperID{3042} 
\def\confName{WACV}
\def\confYear{2026}

\title{CRAUM-Net: Contextual Recursive Attention with Uncertainty Modeling for Salient Object Detection}

\author{Abhinav Sagar\\
University of Maryland, College Park, Maryland\\
College Park, Maryland\\
{\tt\small asagar@umd.edu}
}

\begin{document}
\maketitle

\begin{abstract}

Salient Object Detection (SOD) plays a crucial role in many computer vision applications, requiring accurate localization and precise boundary delineation of salient regions. In this work, we present a novel framework that integrates multi-scale context aggregation, advanced attention mechanisms, and uncertainty-aware module for improved SOD performance. Our Adaptive Cross-Scale Context Module effectively fuses features from multiple levels, leveraging Recursive Channel Spatial Attention and Convolutional Block Attention to enhance salient feature representation. We further introduce an edge-aware decoder that incorporates a dedicated Edge Extractor for boundary refinement, complemented by Monte Carlo Dropout to estimate uncertainty in predictions. To train our network robustly, we employ a combination of boundary-sensitive and topology-preserving loss functions, including Boundary IoU, Focal Tversky, and Topological Saliency losses. Evaluation metrics such as uncertainty-calibrated error and Boundary F1 score, along with the standard SOD metrics, demonstrate our method’s superior ability to produce accurate and reliable saliency maps. Extensive experiments validate the effectiveness of our approach in capturing fine-grained details while quantifying prediction confidence, advancing the state-of-the-art in salient object detection.

\end{abstract}

\section{Introduction}

Salient Object Detection (SOD) focuses on accurately identifying and segmenting the most visually distinctive objects within images. It plays a fundamental role in a wide range of computer vision applications such as image retrieval, object recognition, video analysis, and autonomous systems. Recent advances driven by deep convolutional neural networks have significantly improved the accuracy of saliency prediction. Despite significant progress driven by deep learning, challenges remain in effectively capturing multi-scale context, preserving boundary details, and accurately modeling uncertainty.

\begin{figure}[ht]
\centering
\includegraphics[width=4cm]{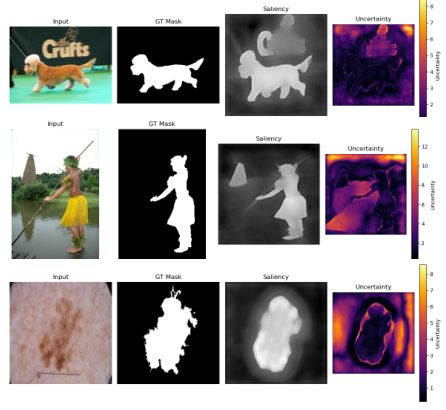}
\caption{The three rows are samples from the datasets DUTS, ECSSD, and ISIC 2018, respectively. From left to right: input image, ground truth, saliency map, and uncertainty map from our method.}
\label{fig:qual_uncertainty_map}
\end{figure}

Uncertainty in SOD arises from multiple sources. Epistemic uncertainty reflects the model’s lack of knowledge and can be reduced with more training data or improved architectures. Aleatoric uncertainty arises from noise and inherent ambiguity in the input data, such as occlusions, complex backgrounds, or low-quality images. Quantifying these uncertainties is crucial for deploying SOD models in real-world, safety-critical scenarios, where understanding the trustworthiness of predictions can prevent costly errors.

The four main contributions of our paper are as follows:

\begin{itemize}
    \item We propose an Adaptive Cross-Scale Context Module that effectively integrates multi-scale features from previous, current, and lateral layers, combined with advanced attention mechanisms including Recursive Channel Spatial Attention and Convolutional Block Attention to enhance salient feature representation and context modeling.
    
    \item We design a hierarchical decoder incorporating boundary refinement via an Edge Extractor module and utilize Monte Carlo Dropout for uncertainty estimation, enabling both precise boundary recovery and reliable uncertainty quantification in saliency maps as shown in Figure~\ref{fig:qual_uncertainty_map}.

    \item Our training framework employs novel loss functions, including boundary-aware and topology-preserving losses, along with novel evaluation metrics, including uncertainty-calibrated error and Boundary F1 score metrics, to ensure robust, accurate, and uncertainty-aware salient object detection.
    
    \item We conduct extensive experiments on standard SOD datasets to validate the effectiveness of our approach and demonstrate that our approach performs better than the state-of-the-art methods, while being interpretable at the same time.
    
\end{itemize}

\section{Related Work}

Salient Object Detection (SOD) has been extensively studied in computer vision to detect the most visually prominent objects in an image. Traditional methods relied on low-level cues such as color contrast, texture, and spatial heuristics \cite{achanta2009frequency,li2014secrets}. However, these handcrafted approaches were limited in handling complex scenes and diverse object appearances.

With the advent of deep learning, CNN-based methods have become the state of the art, exploiting hierarchical feature representations to capture semantic and structural information \cite{liu2019simple,wu2019cascaded,pang2020multi}. Multi-scale feature fusion \cite{wang2020pyramid,hou2017deeply}, attention mechanisms \cite{wang2019salient}, and boundary refinement modules \cite{wu2019cascaded,zhang2019progressive} further improved detection accuracy and localization precision. 

Methods like  Amulet~\cite{zhang2017amulet} introduced multi-level feature fusion and deep supervision, while architectures such as DSS~\cite{hou2017deeply} showed the importance of encoder-decoder structures for dense prediction. Recent works~\cite{liu2019simple} emphasize the importance of boundary precision. By incorporating edge features or boundary-aware losses, these methods produce sharper object contours and improve segmentation accuracy around object borders.

\cite{qin2020u2} introduced U2-Net, a nested U-structure that effectively captures rich multi-scale features to enhance salient object detection with both accuracy and efficiency. Building on the concept of multi-scale contextual modeling, \cite{li2022adjacent} proposed the Adjacent Context Coordination Network (ACCoNet), which facilitates cross-scale feature interaction specifically designed for salient object detection in optical remote sensing images. Complementing these approaches, ~\cite{zhou2024admnet} presented ADMNet, a lightweight and attention-guided densely multi-scale network that balances real-time performance with strong detection capabilities. 

In the context of SOD, uncertainty modeling remains underexplored compared to other vision tasks. Some initial efforts have introduced confidence maps to detect uncertain or ambiguous regions \cite{tian2023modeling, fang2023udnet}, while others have combined uncertainty estimation with multi-task learning \cite{li2021uncertainty}, jointly predicting saliency and boundary uncertainty to improve segmentation quality. However, these approaches often focus on aleatoric uncertainty or employ heuristic confidence measures without principled Bayesian modeling. Moreover, a comprehensive evaluation of uncertainty measures and their practical benefits in SOD is still lacking.

Our approach addresses these gaps by proposing a unified Bayesian framework that explicitly models both epistemic and aleatoric uncertainty in salient object detection. We demonstrate through extensive experiments that uncertainty-aware saliency maps enable more reliable and interpretable predictions, particularly in challenging scenarios involving occlusion, clutter, and low contrast.

\section{Methodology}

\subsection{Problem Definition}

Given an input image $\mathbf{I} \in {R}^{H \times W \times 3}$, the goal of Salient Object Detection (SOD) is to predict a binary saliency map $\mathbf{S} \in \{0,1\}^{H \times W}$ that highlights the most visually prominent objects in the scene. Formally, the task is to learn a mapping function 
\[
f: \mathbf{I} \rightarrow \mathbf{\hat{S}},
\]
where $\mathbf{\hat{S}} \in [0,1]^{H \times W}$ is a probabilistic saliency map representing the likelihood of each pixel belonging to a salient object.

\subsection{Model Architecture}
\label{sec:network}

\subsubsection{Backbone Feature Extraction}

We adopt a modified ResNet50 backbone, denoted as $\text{Backbone\_ResNet50\_in3}$, which extracts feature maps $\{f_1, f_2, f_3, f_4, f_5\}$ at progressively deeper levels. Each feature map $f_i$ is extracted at a spatial resolution of $\frac{1}{2^i}$ of the input image, with increasing channel depth, which are crucial for capturing both global context and fine-grained details necessary for accurate salient object detection. 

\subsubsection{Efficient Channel Attention (ECA) and Spatial Attention (SA) Mechanisms.}

We implement an attention module inspired by the CBAM~\cite{woo2018cbam} framework, combining both channel and spatial attention to enhance feature representations. We replace the channel attention by the CBAM~\cite{woo2018cbam} with Efficient Channel Attention (ECA)~\cite{wang2020eca}. The details of ECA, SA, and CBAM can be found in the Appendix.

\subsubsection{Recursive Channel-Spatial Attention (RCSA).}

To enhance feature representation by iteratively refining both channel and spatial dependencies, we introduce the \textit{Recursive Channel-Spatial Attention (RCSA)} module. The architecture is defined as a recursive attention mechanism that performs joint channel and spatial attention updates over multiple steps, enabling deeper refinement of informative features and suppression of irrelevant activations.

Formally, let $x \in {R}^{C \times H \times W}$ be the input feature map, where $C$ is the number of channels and $H \times W$ denotes the spatial resolution. The module initializes an attention state as $a_0 = x$ and refines it over $T$ steps using both channel and spatial attention cues. Each step involves the following operations:

\begin{enumerate}
    \item \textbf{Depthwise Projection:} A depthwise $1 \times 1$ convolution is applied to retain per-channel semantics while reducing parameter cost:
    \begin{equation}
        a'_t = \text{Conv}_{\text{depth}}(a_{t-1})
    \end{equation}

    \item \textbf{Channel Attention:} We extract channel-wise descriptors using global average pooling and apply a two-layer MLP with reduction ratio $r$ followed by a sigmoid activation:
    \begin{equation}
        \mathbf{c}_t = \sigma\left(W_2 \cdot \text{ReLU}(W_1 \cdot \text{GAP}(a'_t))\right)
    \end{equation}
    where $\sigma$ denotes the sigmoid function, and $W_1$, $W_2$ are $1 \times 1$ convolutional weights with sizes $C \to C/r$ and $C/r \to C$, respectively.

    The attention is then applied to modulate the features:
    \begin{equation}
        a^c_t = a'_t \cdot \mathbf{c}_t
    \end{equation}

    \item \textbf{Spatial Attention:} We compute spatial descriptors by concatenating the channel-wise average and max projections:
    \begin{equation}
        s_t = \sigma\left(\text{Conv}_{7\times7}\left(\left[\text{Avg}(a^c_t); \text{Max}(a^c_t)\right]\right)\right)
    \end{equation}
    where $[\cdot ; \cdot]$ denotes channel-wise concatenation. This generates a spatial mask to refine salient regions.

    \item \textbf{Recursive Update:} The attention state is updated by modulating the previous state with the spatial attention and adding the original residual:
    \begin{equation}
        a_t = a_{t-1} \cdot s_t + x
    \end{equation}
\end{enumerate}

After $T$ recursive steps, the final output is given by:
\begin{equation}
    y = a_T + x
\end{equation}
This residual connection ensures gradient flow and feature preservation.

\subsubsection{Context Aggregation Module.}

To effectively capture multi-scale contextual information, we design a \textit{Context Aggregation Module} (CAM) that leverages parallel dilated convolutions with varying receptive fields. Given an input feature map \( x \in {R}^{B \times C \times H \times W} \), the module applies three parallel \(3 \times 3\) convolutional layers with increasing dilation rates \( \{1, 3, 5\} \). These dilation rates enable the convolutions to sample from local, mid-range, and broader contexts without increasing the number of parameters or the computational burden significantly.

Formally, the output of each branch is computed as:
\[
\begin{aligned}
d_1 &= \text{ReLU}(\text{Conv}_{3\times3}^{\text{dilation}=1}(x)) \\
d_2 &= \text{ReLU}(\text{Conv}_{3\times3}^{\text{dilation}=3}(x)) \\
d_3 &= \text{ReLU}(\text{Conv}_{3\times3}^{\text{dilation}=5}(x))
\end{aligned}
\]

The final output feature map is then obtained by element-wise summation of the three branches:
\[
y = d_1 + d_2 + d_3
\]

\subsubsection{Edge Extractor Module.}

To enhance boundary localization and guide structural learning, we introduce an \textit{Edge Extractor Module} that explicitly captures edge information from input feature maps using Sobel operators. This module computes the image gradients along the horizontal and vertical directions using fixed convolutional filters and then refines the edge response via learnable convolutional layers.

Given an input feature map \( x \in {R}^{B \times 1 \times H \times W} \), horizontal and vertical edges are extracted using classic Sobel kernels:
\[
\text{Sobel}_x = 
\begin{bmatrix}
-1 & 0 & 1 \\
-2 & 0 & 2 \\
-1 & 0 & 1 \\
\end{bmatrix}, \quad
\text{Sobel}_y = 
\begin{bmatrix}
-1 & -2 & -1 \\
0 & 0 & 0 \\
1 & 2 & 1 \\
\end{bmatrix}
\]
These kernels are applied as fixed filters in a convolutional manner:
\[
\text{edge}_x = \text{Conv2D}(x, \text{Sobel}_x), \quad
\text{edge}_y = \text{Conv2D}(x, \text{Sobel}_y)
\]
The gradient magnitude is then computed as:
\[
\text{edges} = \sqrt{\text{edge}_x^2 + \text{edge}_y^2}
\]

To adapt the raw edge maps to the downstream network, we apply a lightweight learnable refinement stage. This consists of a convolutional block (Conv-BN-ReLU) followed by a \(1 \times 1\) convolution and a sigmoid activation:
\[
\text{edges}_{\text{refined}} = \sigma\left( \text{Conv}_{1\times1}(\text{ConvBlock}(\text{edges})) \right)
\]

\subsubsection{Adaptive Cross-Scale Context Module (ACSCM).}

To effectively integrate multi-level features across different stages of the network, we design an Adaptive Cross-scale Context Module based on \cite{li2022adjacent} that aggregates semantic information from three contexts: the \textit{previous}, \textit{current}, and \textit{latter} stages. This module leverages multi-dilated convolutions and attention-based enhancement to form a spatially aware and contextually rich representation.

\textbf{1. Multi-dilated Current Feature Extraction:}  
To model the local and global context from the current stage feature map \( x_{\text{cur}} \), we use four parallel dilated convolution blocks with dilation rates of 1, 2, 3, and 4:
\[
x_{\text{cur}_i} = \text{Conv}_{\text{d}=i}(x_{\text{cur}}), \quad i \in \{1,2,3,4\}
\]
These outputs are concatenated and fused:
\[
x_{\text{cur\_all}} = \text{Conv}_{3\times3}(\text{Concat}[x_{\text{cur}_1}, x_{\text{cur}_2}, x_{\text{cur}_3}, x_{\text{cur}_4}])
\]

\textbf{2. Attention-based Refinement:}  
To further enhance informative regions, the fused current features are passed through a channel-spatial attention module:
\[
x_{\text{cur\_att}} = \text{CBAM}(x_{\text{cur\_all}})
\]

\textbf{3. Previous Feature Encoding:}  
The previous feature map \( x_{\text{pre}} \) is downsampled to match the resolution of the current stage:
\[
x_{\text{pre\_down}} = \text{MaxPool2D}(x_{\text{pre}})
\]
An attention mechanism is then applied:
\[
x_{\text{pre\_att}} = \text{CBAM}(x_{\text{pre\_down}})
\]

\textbf{4. Latter Feature Encoding:}  
The latter stage feature \( x_{\text{lat}} \) is upsampled and enhanced:
\[
x_{\text{lat\_up}} = \text{Upsample}(x_{\text{lat}}), \quad x_{\text{lat\_att}} = \text{CBAM}(x_{\text{lat\_up}})
\]

\textbf{5. Fusion of Multi-scale Context:}  
All attended feature maps are adaptively aligned and added with a residual connection to the current input:
\[
x_{\text{out}} = x_{\text{cur\_att}} + x_{\text{pre\_att}} + x_{\text{lat\_att}} + x_{\text{cur}}
\]

\textbf{Variant - ACSCM1:}  
We adopt a simplified variant, ACSCM1, which omits the latter branch and uses \textit{Recursive Channel-Spatial Attention (RCSA)} instead of CBAM for both current and previous stages.

\subsubsection{Decoder}

To decode multi-level encoder features into a refined prediction map, we design a hierarchical decoder. This module exploits a sequence of dilated and standard convolutions to aggregate contextual information at multiple scales, followed by upsampling and fusion. Each block incorporates the following:

\begin{itemize}
  \item A standard $3 \times 3$ convolution to reduce the input dimensionality.
  \item A dilated convolution (with dilation rates $d_1$ and $d_2$) to enlarge the receptive field without reducing resolution.
  \item Feature fusion using concatenation of intermediate features followed by a final $3 \times 3$ convolution.
\end{itemize}

Formally, for an input feature map $x \in {R}^{C_{in} \times H \times W}$, the block computes:
\begin{align}
    x_1 &= \text{Conv}_{3\times3}(x), \\
    x_1^{d} &= \text{DilatedConv}_{d_1}(x_1), \\
    x_2 &= \text{Conv}_{3\times3}(x_1), \\
    x_2^{d} &= \text{DilatedConv}_{d_2}(x_2), \\
    x_3 &= \text{Conv}_{3\times3}(x_2), \\
    y &= \text{Conv}_{3\times3}([\;x_1^d,\; x_2^d,\; x_3\;]),
\end{align}
where $[\cdot]$ denotes channel-wise concatenation.

\textbf{Progressive Decoding with Skip Fusion.} We design a 5-stage decoder, each stage progressively refining features from coarse-to-fine by:
\begin{enumerate}
  \item Merging upsampled decoder features with the corresponding encoder features via concatenation.
  \item Applying the \texttt{\_Decoder} for local-global context refinement.
  \item Upsampling the output to double its spatial resolution.
  \item Producing intermediate supervision outputs at each stage via $3\times3$ convolution to encourage deep supervision.
\end{enumerate}

Each decoder stage operates as follows:
\begin{align}
    x_i^{\text{up}} &= \text{BAB\_Decoder}([\;x_i,\; \text{Up}(x_{i+1}^{\text{up}})\;]), \\
    s_i &= \text{Conv}_{3\times3}(x_i^{\text{up}}),
\end{align}
where $i \in \{5,4,3,2,1\}$ indexes the decoder stages, and $s_i$ is the output prediction at that stage. Bilinear upsampling is used throughout.

\begin{figure}[ht]
\centering
\includegraphics[width=8cm]{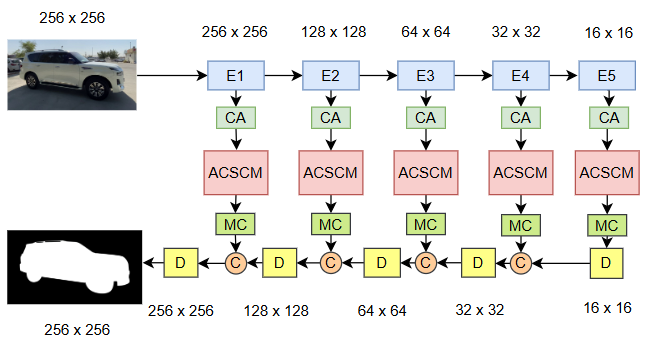}
\caption{Illustration of the proposed network architecture diagram. Here E1, E2, E3, E4, and E5 denote the 5 encoders, CA denotes the Context Aggregation Module, ACSCM denotes the Adaptive Cross-Scale Context Module, MC denotes the Monte-Carlo dropout, D denotes the decoder, and C denotes the concatenation operation.}
\label{fig:model_architecture}
\end{figure}

\subsubsection{Uncertainty Quantification in Salient Object Detection}

To effectively quantify uncertainty in our SOD framework, we employ Monte Carlo (MC) Dropout \cite{gal2016dropout} as a practical Bayesian approximation. This approach allows us to estimate predictive uncertainty by performing multiple stochastic forward passes through the network at inference time, capturing model uncertainty caused by limited training data or model capacity.

Given an input image $\mathbf{x}$, the model outputs a saliency prediction $\mathbf{\hat{y}}$ by forwarding through the network with dropout enabled during inference. We perform $T$ stochastic forward passes, obtaining a set of predictions $\{\mathbf{\hat{y}}^{(t)}\}_{t=1}^T$. The final saliency map $\mathbf{\hat{y}}$ is estimated as the expectation:

\begin{equation}
\mathbf{\hat{y}} = \frac{1}{T} \sum_{t=1}^T \mathbf{\hat{y}}^{(t)}
\end{equation}

The predictive uncertainty is quantified by the variance of these predictions, reflecting the model's confidence:

\begin{equation}
\mathbf{U} = \frac{1}{T} \sum_{t=1}^T \left(\mathbf{\hat{y}}^{(t)} - \mathbf{\hat{y}}\right)^2
\end{equation}

Here, $\mathbf{U}$ is the pixel-wise uncertainty map, with higher values indicating regions where the model is less confident.

\subsubsection{Final Prediction}

The final prediction is generated by projecting the highest-resolution fused feature map to a single channel:

\begin{equation}
    S = \sigma(\text{Conv}_{1 \times 1}(F_1))
\end{equation}

This output $S$ represents the saliency map. Figure~\ref{fig:model_architecture} illustrates the overall architecture of our proposed framework.

\subsection{Loss Function}

SOD demands not only accurate pixel-wise classification but also structural and topological fidelity. Traditional loss functions like Binary Cross-Entropy (BCE) may result in blurry or boundary-inaccurate predictions. To overcome these limitations, we propose a composite loss function that jointly optimizes pixel-level accuracy, overlap quality, boundary precision, and topological consistency.

\paragraph{1. Binary Cross-Entropy (BCE) Loss:}. The BCE loss is referred as $\mathcal{L}_{\text{BCE}}$. The details can be found in the Appendix. 

\paragraph{2. IoU Loss:} The IoU loss is referred as $\mathcal{L}_{\text{IoU}}$. The details can be found in the Appendix. 

\paragraph{3. Focal Tversky (FT) Loss:} The FT loss is referred as $\mathcal{L}_{\text{FT}}$. The details can be found in the Appendix. 

\paragraph{4. Boundary IoU (BIoU) Loss:}
To sharpen object contours and improve boundary localization, we introduce the Boundary IoU loss. Unlike pixel-level or region-level losses, BIoU operates on edge maps derived via gradient approximations:

\begin{align}
\mathcal{L}_{\text{BIoU}} = 1 - \frac{ \sum (E_P \cdot E_G) }{ \sum E_P + \sum E_G - \sum (E_P \cdot E_G) + \epsilon }
\end{align}

where $E_P$ and $E_G$ denote the predicted and ground truth edge maps, respectively, computed using horizontal and vertical edge differences. 

\paragraph{5. Topological Saliency (TS) Loss:}
Salient objects are often expected to be structurally coherent (e.g., contiguous, hole-free regions). To preserve topological consistency, we present a Laplacian-based loss that compares structural contours:

\begin{align}
\mathcal{L}_{\text{TS}} = \| \nabla^2(P > 0.5) - \nabla^2(G > 0.5) \|_1
\end{align}

where $P$ is the predicted probability map and $G$ is the ground-truth binary map, $\nabla^2$ is the discrete Laplacian operator, and $(\cdot > 0.5)$ is the binarization threshold. 

Our final loss function is defined as:

\begin{align}
\mathcal{L}_{\text{total}} = \lambda_1 \mathcal{L}_{\text{BCE}} + \lambda_2 \mathcal{L}_{\text{IoU}} + \lambda_3 \mathcal{L}_{\text{FT}} + \lambda_4 \mathcal{L}_{\text{BIoU}} + \lambda_5 \mathcal{L}_{\text{TS}}
\end{align}

where $\lambda_i$ are weights empirically set to balance the contribution of each component. In our experiments, we empirically set the weights as:
\[
\lambda_1 = 1.0, \quad \lambda_2 = 1.0, \quad \lambda_3 = 0.8, \quad \lambda_4 = 0.5, \quad \lambda_5 = 0.3
\]

\section{Experiments}

\subsection{Dataset}

The following datasets were used to test and compare the performance of our method:

\textbf{DUTS, ECSSD, HKU-IS, SBU-Shadow, and ISIC2018.} The details can be found in the Appendix. 

\subsection{Implementation Details}

We implemented our SOD framework using the PyTorch deep learning library. All experiments were conducted on a workstation equipped with an Nvidia A100 GPU, an Intel Core i9 CPU, and 128 GB RAM, running Ubuntu 20.04 and Python 3.10. All input images and corresponding ground truth saliency maps were resized to $256 \times 256$ before feeding into the network. We used the Adam optimizer with an initial learning rate of $1 \times 10^{-4}$ and weight decay of $5 \times 10^{-4}$. The learning rate was scheduled to decay using a polynomial annealing policy. Models were trained for 100 epochs with a batch size of 4. Gradient clipping was applied with a threshold of 1.0 to stabilize training. During testing, only resizing and normalization were applied. The final saliency maps were upsampled to the original image resolution using bilinear interpolation.

\subsection{Evaluation Metrics}

To comprehensively assess the performance of the SOD models, we employ a suite of widely adopted evaluation metrics. These metrics quantify different aspects of prediction quality, including pixel-level accuracy, structural consistency, and region-level overlap with ground truth annotations.

\paragraph{Mean Absolute Error (MAE), F-measure $\mathbf{F_{max}}$, $\mathbf{F_{adaptive}}$, and $\mathbf{F_{weighted}}$, Structure Measure ($S_\alpha$), Enhanced-alignment Measure ($E_\phi$), and Intersection over Union (IoU).} The details can be found in the Appendix. 

We propose 2 new evaluation metrics for SOD: Uncertainty-Calibrated Error and Boundary-aware F-measure, as explained below:

\paragraph{Uncertainty-Calibrated Error (UCE).}
Traditional saliency metrics such as MAE or F-measure evaluate the correctness of predictions but do not account for the model's confidence in its outputs. The Uncertainty-Calibrated Error (UCE) bridges this gap by integrating the model’s predictive uncertainty into error evaluation. It penalizes predictions that are both wrong and overconfident, aligning with the principle that a well-calibrated model should be uncertain when it is likely to err. Formally, for a probabilistic prediction $\hat{Y} \in [0,1]$ and a ground truth mask $Y \in \{0,1\}$, UCE is defined as:

\begin{equation}
\text{UCE} = {E} \left[ \left(1 - \mathcal{H}(\hat{Y})\right) \cdot {1}_{|\hat{Y} - Y| > \delta} \right],
\end{equation}

where $\delta$ is a predefined error tolerance ($\delta = 0.5$), and $\mathcal{H}(\hat{Y})$ is the binary entropy of the prediction:

\begin{equation}
\mathcal{H}(\hat{Y}) = -\hat{Y} \log \hat{Y} - (1 - \hat{Y}) \log (1 - \hat{Y}).
\end{equation}

The UCE term $(1 - \mathcal{H}(\hat{Y}))$ penalizes confident predictions (low entropy), while ${1}_{|\hat{Y} - Y| > \delta}$ ensures that only significantly incorrect predictions are considered. 

Intuitively, UCE is high when the model makes wrong predictions with high confidence, and low when errors are accompanied by high uncertainty. UCE rewards uncertainty in ambiguous regions (e.g., edges, occlusions). High UCE would correlate with robustness failures in safety-critical or low-data regimes. In practice, UCE is especially relevant in SOD tasks involving ambiguous foreground-background transitions. Lower UCE values indicate better error-uncertainty alignment and improved trustworthiness in model predictions.

\paragraph{Boundary-aware F-measure (B-F1).}

While traditional F-measure evaluates saliency prediction at the pixel level, it does not explicitly consider the spatial quality of object boundaries. The Boundary-aware F-measure (B-F1) addresses this by evaluating how well the predicted and ground truth boundaries align. First, the boundary maps of both the predicted saliency mask $P$ and the ground truth $Y$ are extracted using morphological gradient operations (dilation minus erosion). A tolerance-based matching is then performed using a disk-shaped structuring element of a given radius (2 pixels), allowing for slight misalignments between boundary pixels.

\begin{equation}
\text{BF1} = \frac{2 \cdot \text{Precision}_b \cdot \text{Recall}_b}{\text{Precision}_b + \text{Recall}_b},
\end{equation}

where

\begin{equation}
\begin{aligned}
\text{Precision}_b &= \frac{|\text{PB} \cap \text{DGTB}|}{|\text{PB}|}, \\
\text{Recall}_b &= \frac{|\text{GTB} \cap \text{DPB}|}{|\text{GTB}|}.
\end{aligned}
\end{equation}

Here, PB denotes Predicted Boundary, DGTB denotes Dilated Ground Truth Boundary, GTB denotes Ground Truth Boundary, and DPB denotes Dilated Predicted Boundary, respectively. The dilation accounts for localization tolerance, ensuring that near-boundary matches are rewarded. A higher B-F1 indicates sharper, more accurate boundary predictions.

\paragraph{Computational Efficiency Metrics.}

Number of Parameters (Params) in Millions, Floating Point Operations (FLOPs), Inference Time in milliseconds (ms), and Frames Per Second (FPS). The details can be found in the Appendix. All computational metrics are measured under standardized conditions: batch size of 1, fixed input resolution (256$\times$256), and on the same hardware. 

\subsection{Performance Comparisons}

We compare our model with other state-of-the-art salient object detection methods. The performance of these methods is reproduced using open-sourced implementations provided by the respective authors, under similar experimental settings. All models are evaluated on the same testing datasets and with the same preprocessing and resolution configurations.

\subsubsection{Quantitative Performance}

We evaluate the models in terms of performance and computational complexity. Table~\ref{tab:quant_results} summarizes the quantitative results on multiple benchmark datasets, including DUTS-TE, ECSSD, HKU-IS, SBU-Shadow, and the ISIC 2018 dataset with other state-of-the-art methods.

\begin{table*}[t]
\tiny
\centering
\caption{A: Quantitative comparison of our approach with other state-of-the-art methods on the following datasets: A: DUTS, B: ECSSD, C: HKU-IS, D: SBU-Shadow, E: ISIC 2018. The best values are highlighted in blue.}
\label{tab:quant_results}
\resizebox{\linewidth}{!}{
\begin{tabular}{l|cccccccccccccc}
\toprule

& \multicolumn{9}{c}{A: DUTS Dataset}  \\ 

\hline
\textbf{Method} & MAE $\downarrow$ & $F(m)$ $\uparrow$ & $F(a)$ $\uparrow$ & $F(w)$ $\uparrow$ &
$S$ $\uparrow$ & $E$ $\uparrow$ & IoU $\uparrow$ & UCE $\downarrow$ & B-F1 $\uparrow$ & Params $\downarrow$ & Flops $\downarrow$ & Time $\downarrow$& FPS $\uparrow$\\
\midrule
ACCoNet \cite{li2022adjacent}       & 0.0544& 0.8792& 0.7705& 0.7947& \textcolor{blue}{0.8570}& 0.9248& 0.7209& 0.2325& 0.6460& 127.01& 51.33& 0.0179& 55.96\\
ADMNet \cite{zhou2024admnet}      & 0.0915& 0.7794& 0.6490& 0.6803& 0.7721& 0.8719& 0.5847& 0.2515& 0.4925& 0.84& 0.42& 0.0293& 34.16\\
BBRF \cite{ma2023boosting}        & \textcolor{blue}{0.0486}& 0.8462& \textcolor{blue}{0.7848}& 0.7824& 0.8476 & 0.8714& 0.7023& \textcolor{blue}{0.0751}& 0.6594& 74.40& 27.12& 0.1200& 8.33\\
C2FNet \cite{sun2021context}      & 0.0619& 0.8680& 0.6909& 0.7923& 0.8459& 0.9249& 0.7119& 0.2528& 0.6124& 28.41& 6.96& 0.0253&39.46\\
CorrNet \cite{gongyangli2022lightweight} &0.0925 & 0.7877& 0.7216& 0.5358& 0.6932& 0.8335& 0.4371& 0.2005& 0.4293& 4.09& 21.31& \textcolor{blue}{0.0094}& \textcolor{blue}{106.91}\\
MINet \cite{shen2024minet}       & 0.0951& 0.7781& 0.6282& 0.6477& 0.7593& 0.8857& 0.5583
& 0.2537& 0.4397& \textcolor{blue}{0.28}& \textcolor{blue}{0.16}& 0.2152&4.65\\
SOC \cite{fan2022salient} & 0.0522& \textcolor{blue}{0.8946}& 0.7566& 0.7616& 0.8503& \textcolor{blue}{0.9491}& 0.6964 & 0.2872& 0.5678& 100.01& 100.98& 0.0107& 93.45\\
U2Net \cite{qin2020u2} & 0.0653& 0.8343& 0.7171& 0.7549& 0.8274& 0.8961& 0.6736& 0.2168& 0.6098& 44.01& 37.65& 0.0395&25.34\\
Ours & 0.0502& 0.8597& 0.7780& \textcolor{blue}{0.8042}& 0.8566& 0.9041& \textcolor{blue}{0.7247}& 0.2036& \textcolor{blue}{0.6901}& 153.57& 61.56& 0.1030& 9.71\\

\hline

& \multicolumn{9}{c}{B: ECSSD Dataset}  \\ 

\hline

ACCoNet \cite{li2022adjacent}       & 0.0409& \textcolor{blue}{0.9597}& \textcolor{blue}{0.8838}& 0.9033& \textcolor{blue}{0.9142}& \textcolor{blue}{0.9514}& \textcolor{blue}{0.8556}& 0.2916& 0.7307& 127.01& 51.33& 0.0177& 56.58\\
ADMNet \cite{zhou2024admnet}      & 0.0755& 0.9100& 0.8207& 0.8182& 0.8433& 0.8911& 0.7417& 0.2985& 0.5931& 0.84& 0.42& 0.0293& 34.16\\
BBRF \cite{ma2023boosting}        & 0.0433& 0.9351& 0.8764& 0.8899& 0.8972& 0.9170& 0.8333& \textcolor{blue}{0.0615}& 0.7608& 74.40& 27.12& 0.1436& 6.96\\
C2FNet \cite{sun2021context}      & 0.0453& 0.9461& 0.8187& 0.8975& 0.9034& 0.9470& 0.8429& 0.2998& 0.6999& 28.41& 6.96& 0.0301&33.20\\
CorrNet \cite{gongyangli2022lightweight} &0.1302 & 0.8720& 0.8277& 0.5968& 0.6963& 0.7827& 0.5001& 0.2671& 0.4446& 4.09& 21.31& 0.0093&107.02\\
MINet \cite{shen2024minet} & 0.0845& 0.9088& 0.8006& 0.7936& 0.8343& 0.8859& 0.7238& 0.3048& 0.5351& \textcolor{blue}{0.28}& \textcolor{blue}{0.16}& 0.1896&5.27\\
SOC \cite{fan2022salient}       & 0.0543& 0.9579& 0.8791& 0.8608& 0.8914& 0.9473& 0.8113& 0.3390& 0.6291& 100.01& 100.98& \textcolor{blue}{0.0084}&\textcolor{blue}{119.00}\\
U2Net \cite{qin2020u2}       & 0.0535& 0.9411& 0.8628& 0.8739& 0.8895& 0.9228& 0.8151& 0.2535& 0.7111& 44.01& 37.65& 0.0419&23.89\\
Ours & \textcolor{blue}{0.0408}& 0.9495& 0.8754& \textcolor{blue}{0.9041}& 0.9067& 0.9395& 0.8500& 0.2693& \textcolor{blue}{0.7848}& 153.57&61.56 & 0.1055&9.48\\

\hline

& \multicolumn{9}{c}{C: HKU-IS Dataset}  \\ 

\hline

ACCoNet \cite{li2022adjacent}       & 0.0365& \textcolor{blue}{0.9481}& 0.8963& 0.8831& \textcolor{blue}{0.9049}& 0.9540& 0.8264& 0.2463& 0.7336& 127.01& 51.33& 0.0214& 46.70\\
ADMNet \cite{zhou2024admnet}      & 0.0673& 0.8902& 0.8014& 0.7982& 0.8371& 0.9080& 0.7147& 0.2638& 0.6014& 0.84& 0.42& 0.0258& 38.71\\
BBRF \cite{ma2023boosting}        & 0.0391& 0.9208& 0.8898& 0.8638& 0.8826 & 0.9180& 0.7952& \textcolor{blue}{0.0558}& 0.7440& 74.40& 27.12& 0.1221& 8.19\\
C2FNet \cite{sun2021context}      & 0.0418& 0.9359& 0.8361& 0.8788&0.8960 & 0.9518& 0.8162& 0.2561& 0.6955& 28.41& 6.96& 0.0242&41.33\\
CorrNet \cite{gongyangli2022lightweight} & 0.0955& 0.8726& 0.8366& 0.6304& 0.7268& 0.8352& 0.5267& 0.2244& 0.4993& 4.09& 21.31& 0.0114&87.92\\
MINet \cite{shen2024minet}       & 0.0750& 0.8883& 0.7802& 0.7698& 0.8259& 0.9054& 0.6908& 0.2695& 0.5422& \textcolor{blue}{0.28}& \textcolor{blue}{0.16}& 0.1964&5.09\\
SOC \cite{fan2022salient}       & 0.0443& 0.9454& 0.8774& 0.8467& 0.8897& \textcolor{blue}{0.9557}& 0.7931& 0.3016& 0.6476& 100.01& 100.98 & \textcolor{blue}{0.0112}&\textcolor{blue}{89.58}\\
U2Net \cite{qin2020u2}       & 0.0451& 0.9261& 0.8580& 0.8622& 0.8876& 0.9365& 0.7995& 0.2217& 0.7144& 44.01& 37.65& 0.0312&32.05\\
Ours & \textcolor{blue}{0.0334}& 0.9371& \textcolor{blue}{0.8969}& \textcolor{blue}{0.8931}& 0.9041& 0.9474& \textcolor{blue}{0.8334}& 0.2231& \textcolor{blue}{0.7802}& 153.57& 61.56& 0.1000&10.00\\

\hline

& \multicolumn{9}{c}{D: SBU-Shadow Dataset}  \\ 

\hline

ACCoNet \cite{li2022adjacent}       & 0.0360& 0.9293& 0.8051& 0.8635& 0.8713& 0.9476& 0.7947& 0.2520& 0.6899& 127.01& 51.33& 0.0184& 54.49\\
ADMNet \cite{zhou2024admnet}      & 0.1288& 0.7594& 0.6898& 0.4798& 0.6295& 0.7581& 0.3835& 0.2996& 0.3473& 0.84& 0.42& 0.0246& 40.60\\
BBRF \cite{ma2023boosting}        & 0.0338& 0.9216& \textcolor{blue}{0.8447}& 0.8682& 0.8696& 0.9256& 0.7995& \textcolor{blue}{0.0502}& \textcolor{blue}{0.7404}& 74.40& 27.12& 0.0754&13.27\\
C2FNet \cite{sun2021context}      & 0.0387& 0.9209& 0.7978& 0.8359& 0.8593& 0.9381& 0.7672& 0.2390& 0.6372& 28.41& 6.96& 0.0213&46.87\\
CorrNet \cite{gongyangli2022lightweight} & 0.0571& 0.9127& 0.8263& 0.7808& 0.8160& 0.9294& 0.6972& 0.2429& 0.5953& 4.09& 21.31&\textcolor{blue}{0.0111} & \textcolor{blue}{90.42}\\
MINet \cite{shen2024minet} & 0.1497& 0.7327& 0.6427& 0.3870& 0.5836& 0.7927& 0.3136& 0.2823& 0.2518& \textcolor{blue}{0.28}& \textcolor{blue}{0.16}& 0.1918&5.21\\
SOC \cite{fan2022salient}       & 0.0444& \textcolor{blue}{0.9347}& 0.8087& 0.8183& 0.8530& \textcolor{blue}{0.9527}& 0.7556& 0.2771& 0.5924& 100.01& 100.98& 0.0119&84.21\\
U2Net \cite{qin2020u2}       & 0.0810& 0.7955& 0.6966& 0.6927& 0.7594& 0.8167& 0.6197& 0.2322& 0.5516& 44.01& 37.65& 0.0642&15.58\\
Ours & \textcolor{blue}{0.0316}& 0.9303& 0.8386& \textcolor{blue}{0.8736}& \textcolor{blue}{0.8771}& 0.9457& \textcolor{blue}{0.8053}& 0.2238& 0.7396& 153.57& 61.56& 0.0991&10.09\\

\hline

& \multicolumn{9}{c}{E: ISIC Dataset}  \\ 

\hline

ACCoNet \cite{li2022adjacent}       & 0.0557& 0.9265& 0.7394& 0.8815& 0.8770& 0.9142& 0.8050& 0.2891& 0.4896& 127.01& 51.33& 0.0182& 54.84\\
ADMNet \cite{zhou2024admnet}      & 0.0639& 0.9356& 0.7735& 0.8602& 0.8684& 0.8939& 0.7844& 0.2845& 0.4815&0.84& 0.42& 0.0246& 40.61\\
BBRF \cite{ma2023boosting}        & 0.0575& 0.9086& 0.7602& 0.8705& 0.8727& 0.8933& 0.7931& \textcolor{blue}{0.0855}& 0.4888&74.40& 27.12& 0.0692&14.45\\
C2FNet \cite{sun2021context}      & 0.0582& 0.9196& 0.7302& 0.8791& 0.8753& 0.9159& 0.8028& 0.2931& 0.4797&28.41& 6.96& 0.0208&48.01\\
CorrNet \cite{gongyangli2022lightweight} &0.1406 & 0.8981& \textcolor{blue}{0.8387}& 0.6509&0.7320 & 0.7661& 0.5655& 0.2870& 0.3975&4.09& 21.31& 0.0116&86.57\\
MINet \cite{shen2024minet} & 0.0627& 0.9583& 0.7420& 0.8631& 0.8770& 0.9186& 0.7947& 0.3027& 0.4878&\textcolor{blue}{0.28}& \textcolor{blue}{0.16}& 0.1856&5.39\\
SOC \cite{fan2022salient}       & 0.0601& \textcolor{blue}{0.9618}& 0.6893& 0.8589& \textcolor{blue}{0.8827}& \textcolor{blue}{0.9459}& 0.7947& 0.3847& 0.4264&100.01& 100.98& \textcolor{blue}{0.0085}& \textcolor{blue}{117.93}\\
U2Net \cite{qin2020u2}       & 0.0623& 0.9553& 0.7522& 0.8626& 0.8734& 0.9130& 0.7879& 0.1831& 0.4685&44.01& 37.65& 0.0475&21.07\\
Ours & \textcolor{blue}{0.0550}& 0.9055& 0.7349& \textcolor{blue}{0.8825}& 0.8752& 0.9034& \textcolor{blue}{0.8075}& 0.2875& \textcolor{blue}{0.4981}& 153.57& 61.56& 0.1022&9.78\\

\bottomrule
\end{tabular}
}
\end{table*}

Our proposed method outperforms existing models across multiple metrics and datasets, demonstrating its superior ability to detect salient objects.

\subsubsection{Qualitative Performance}

\begin{figure}[ht]
\centering
\includegraphics[width=8cm]{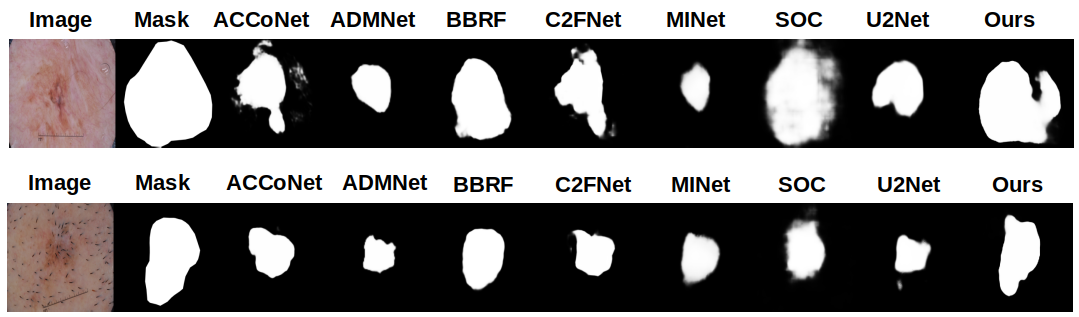}
\caption{Qualitative comparison of different methods on the ISIC 2018 dataset. From left to right: input image, ground truth, ACCoNet, ADMNet, BBRF, C2FNet, MINet, SOC, U2Net, and our method.}
\label{fig:qual_results_isic}
\end{figure}

\begin{figure}[ht]
\centering
\includegraphics[width=8cm]{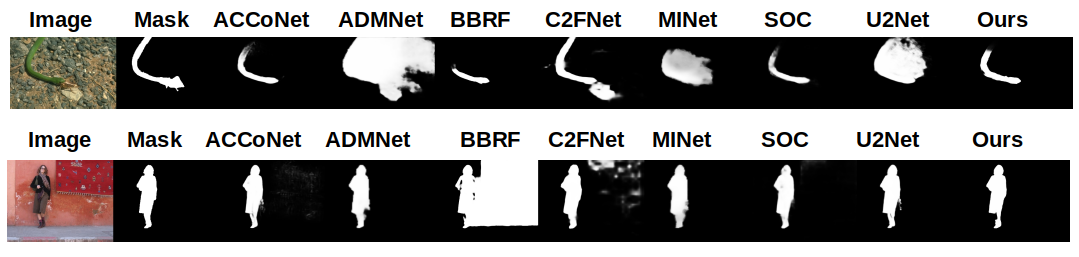}
\caption{Qualitative comparison of different methods on the ECSSD dataset. From left to right: input image, ground truth, ACCoNet, ADMNet, BBRF, C2FNet, MINet, SOC, U2Net, and our method.}
\label{fig:qual_results_ecssd}
\end{figure}

\begin{figure}[ht]
\centering
\includegraphics[width=8cm]{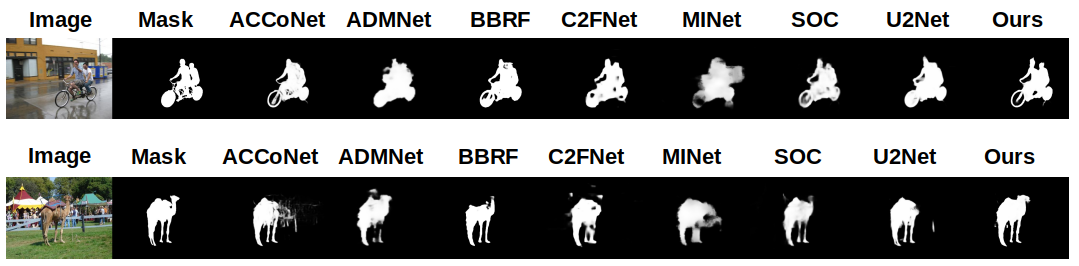}
\caption{Qualitative comparison of different methods on the DUTS dataset. From left to right: input image, ground truth, ACCoNet, ADMNet, BBRF, C2FNet, MINet, SOC, U2Net, and our method.}
\label{fig:qual_results_duts}
\end{figure}

Figure~\ref {fig:qual_results_isic}, ~\ref {fig:qual_results_ecssd} and ~\ref{fig:qual_results_duts} show visual comparisons of our method against several baselines. Our model accurately delineates salient regions, preserves boundary structures, and avoids false positives in complex backgrounds.

Overall, these results confirm the effectiveness of incorporating uncertainty estimation into the SOD pipeline, improving both prediction quality and model trustworthiness.

\subsection{Ablation Study}

To rigorously evaluate the contribution of each proposed component in our uncertainty-aware salient object detection framework, we conduct comprehensive ablation studies using the DUTS dataset. We individually remove modules from the baseline to isolate and quantify their individual and joint effects.

\begin{table*}[t]
\tiny
\centering
\caption{Ablation study on DUTS dataset by removing components. $\downarrow$: lower is better; $\uparrow$: higher is better. Best results in \textcolor{blue}{blue}.}
\label{tab:ablation_network}
\begin{tabular}{l|cccccccccccc}
\toprule
\textbf{Method} & MAE$\downarrow$ & $F_m$$\uparrow$ & $F_a$$\uparrow$ & $F_w$$\uparrow$ & $S$$\uparrow$ & $E$$\uparrow$ & IoU$\uparrow$ & UCE$\downarrow$ & B-F1$\uparrow$ & Param$\downarrow$ & FLOPs$\downarrow$ & FPS$\uparrow$ \\
\midrule
W/o CA     & 0.0534 & 0.8584 & 0.7727 & 0.8054 & 0.8540 & 0.8973 & 0.7236 & 0.2052 & \textcolor{blue}{0.6946} & \textcolor{blue}{137.08} & \textcolor{blue}{53.86} & 11.29 \\
W/o CBAM   & 0.0506 & 0.8561 & 0.7753 & 0.8008 & 0.8558 & 0.8981 & 0.7211 & \textcolor{blue}{0.1976} & 0.6868 & 153.54 & 61.54 & 10.44 \\
W/o DropP. & 0.0530 & 0.8580 & 0.7630 & 0.7949 & 0.8541 & 0.9034 & 0.7188 & 0.2490 & 0.6605 & 153.57 & 61.56 & 10.10 \\
W/o Edge   & 0.0516 & 0.8570 & 0.7749 & \textcolor{blue}{0.8060} & 0.8551 & 0.9014 & 0.7240 & 0.2016 & 0.6920 & 153.57 & 61.54 & 10.74 \\
W/o MC Drop& 0.0516 & 0.8582 & 0.7744 & 0.8028 & 0.8562 & 0.9001 & 0.7238 & 0.1989 & 0.6922 & 153.57 & 61.56 & 10.16 \\
W/o RecAtt & 0.0507 & 0.8565 & 0.7720 & 0.8009 & 0.8559 & 0.8910 & 0.7215 & 0.2007 & 0.6875 & 148.40 & 61.49 & 11.97 \\
\textbf{Ours}     & 0.0502 & \textcolor{blue}{0.8597} & \textcolor{blue}{0.7780} & 0.8042 & 0.8566 & \textcolor{blue}{0.9041} & \textcolor{blue}{0.7247} & 0.2036 & 0.6901 & 153.57 & 61.56 & 9.71 \\
\bottomrule
\end{tabular}
\end{table*}

\paragraph{Quantitative Results.} Table~\ref{tab:ablation_network} presents a comprehensive ablation study on the DUTS dataset to evaluate the impact of each component in our network architecture. We systematically remove one module at a time while keeping the rest of the architecture unchanged. The full model (denoted as "Ours") achieves the best overall performance in terms of F-measure variants, enhanced alignment ($E$), IoU, and overall structural quality, validating the synergy of all components. 


\begin{table}[t]
\centering
\caption{Ablation study on DUTS dataset by removing loss components. $\downarrow$ lower is better; $\uparrow$ higher is better. Best values are in blue.}
\label{tab:ablation_loss}
\resizebox{\columnwidth}{!}{%
\begin{tabular}{l|ccccccccc}
\toprule
\textbf{Method} & MAE$\downarrow$ & \makecell{$F_m$\\$\uparrow$} & \makecell{$F_a$\\$\uparrow$} & \makecell{$F_w$\\$\uparrow$} & S$\uparrow$ & E$\uparrow$ & IoU$\uparrow$ & \makecell{UCE\\$\downarrow$} & \makecell{B-F1\\$\uparrow$} \\
\midrule
W/o B-IOU & \textcolor{blue}{0.0484} & 0.8512 & 0.7769 & 0.8040 & \textcolor{blue}{0.8606} & 0.9074 & 0.7163 & 0.2033 & 0.6792 \\
W/o CE    & 0.0533 & 0.8206 & 0.7756 & 0.7973 & 0.8500 & 0.8506 & 0.7150 & 0.2087 & 0.6832 \\
W/o FT    & 0.0537 & 0.8544 & 0.7701 & 0.7989 & 0.8535 & \textcolor{blue}{0.9181} & 0.7206 & 0.2076 & 0.6846 \\
W/o IOU   & 0.0495 & 0.8529 & 0.7754 & 0.7980 & 0.8556 & 0.9077 & 0.7189 & 0.2072 & 0.6804 \\
W/o TS    & 0.0524 & 0.8518 & 0.7767 & 0.8003 & 0.8529 & 0.8820 & 0.7218 & \textcolor{blue}{0.2003} & 0.6885 \\
Ours      & 0.0502 & \textcolor{blue}{0.8597} & \textcolor{blue}{0.7780} & \textcolor{blue}{0.8042} & 0.8566 & 0.9041 & \textcolor{blue}{0.7247} & 0.2036 & \textcolor{blue}{0.6901} \\
\bottomrule
\end{tabular}
}
\end{table}

Table~\ref{tab:ablation_loss} presents an ablation study assessing the impact of each individual loss function on the overall performance of our model. We systematically remove one loss term at a time from our full loss formulation and evaluate the model across several metrics. Overall, the full model—with all loss components combined—achieves the best balance across all performance metrics, thereby validating the complementary effects of our multi-term loss design.


\paragraph{Qualitative Analysis.} 

\begin{figure}[ht]
\centering
\includegraphics[width=6cm]{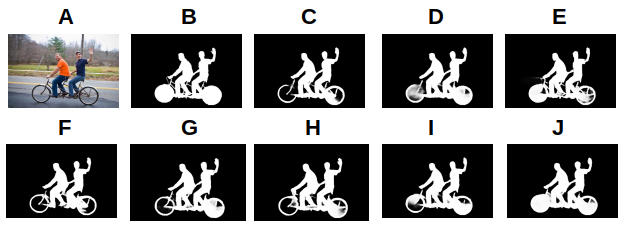}
\caption{Qualitative ablation study by removing individual components from the network architecture using the DUTS dataset. A: Image, B: Ground Truth, C: Without CA Module, D: Without CBAM Attention, E: Without DropPath, F: Without Edge Extractor, G: Without MC Dropout, H: Without Recursive Attention, J: Our Model.}
\label{fig:ablation_network}
\end{figure}

Figure~\ref{fig:ablation_network} visually illustrates the impact of removing individual architectural components on the saliency prediction quality, using samples from the DUTS dataset. Compared to the ground truth (B), the outputs from ablated models (C--I) exhibit various forms of degradation. In contrast, our full model (J) produces sharp, coherent, and spatially aligned saliency maps, affirming the synergistic contribution of each architectural component toward robust saliency detection.


\begin{figure}[ht]
\centering
\includegraphics[width=8cm]{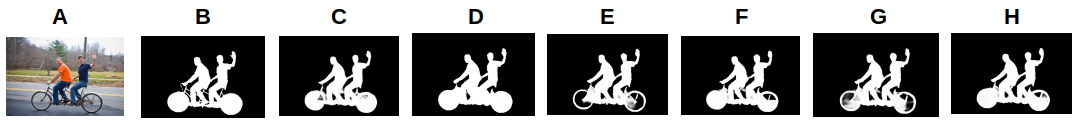}
\caption{Qualitative ablation study by removing individual loss functions using the DUTS dataset. A: Image, B: Ground-Truth, C: Without B-IOU loss, D: Without CE loss, E: Without Focal Tversky loss, F: Without IOU loss, G: Without Topological Saliency loss, H: Our loss.}
\label{fig:ablation_loss}
\end{figure}

Figure~\ref{fig:ablation_loss} presents a qualitative comparison of saliency maps generated by ablated models trained without specific loss components, using the DUTS dataset. Each column shows the effect of removing one individual loss term. The full model trained with all loss components (H) produces sharp, well-aligned, and semantically coherent saliency maps, underscoring the complementary strengths of each loss term in optimizing both pixel-level accuracy and structural fidelity.


\section{Conclusions}

We propose a novel salient object detection framework that fuses multi-scale context, advanced attention, and uncertainty modeling to improve accuracy and boundary precision. Our model combines Recursive Channel Spatial Attention, Adaptive Cross-Scale Context modules, and an Edge Extractor to capture rich context and fine details, while Monte Carlo Dropout provides reliable uncertainty estimation. Experiments show superior performance on standard benchmarks, and boundary-sensitive, topology-preserving losses further enhance the precision and consistency of saliency maps.

\nocite{*}
{
    \small
    \bibliographystyle{ieeenat_fullname}
    \bibliography{main}
}

\newpage

\section{Supplementary Material}

We present additional details of our work in the following sections:

\subsection{Network Architecture}

\subsubsection{Efficient Channel Attention (ECA) and Spatial Attention Mechanisms.}

\textbf{Efficient Channel Attention (ECA)} focuses on capturing channel-wise dependencies without dimensionality reduction or excessive parameter overhead. Given an input feature map \( x \in {R}^{B \times C \times H \times W} \), ECA applies global average pooling to squeeze the spatial dimensions:

\[
z = \text{GAP}(x) \in {R}^{B \times C \times 1 \times 1}
\]

The pooled vector is reshaped and permuted to \( {R}^{B \times 1 \times C} \) to prepare for a 1D convolution along the channel dimension. This convolution with kernel size \( k \) models local cross-channel interactions efficiently:

\[
z' = \sigma \big(\text{Conv1D}(z^\top)\big) \in {R}^{B \times C \times 1}
\]

where \( \sigma \) denotes the sigmoid activation. After restoring the tensor shape to \( {R}^{B \times C \times 1 \times 1} \), the attention weights modulate the original input feature map by element-wise multiplication, effectively re-weighting each channel.

\textbf{Spatial Attention (SA)} captures spatial importance by aggregating channel information through two pooling operations: max-pooling and average-pooling along the channel axis, producing two spatial descriptors:

\[
M_{\text{max}} = \max_{c}(x), \quad M_{\text{avg}} = \text{mean}_{c}(x)
\]

These descriptors are concatenated and passed through a convolutional layer with kernel size \( k \) followed by a sigmoid activation, generating a spatial attention mask:

\[
M_s = \sigma \big( \text{Conv}_{k \times k}([M_{\text{max}}; M_{\text{avg}}]) \big)
\]

This mask is then applied to the input via element-wise multiplication, highlighting the spatially important regions.

\textbf{Combined CBAM Block.} The final CBAM block sequentially applies ECA and SA attention modules on the input feature map. Specifically, the input \( x \) is first channel-refined by ECA, then spatially refined by SA. A residual connection adds the original input to the output, preserving low-level features and stabilizing training:

\[
\begin{aligned}
x' &= x \times \text{ECA}(x) \\
x'' &= x' \times \text{SA}(x') \\
\text{Output} &= x'' + x
\end{aligned}
\]

\subsubsection{Adaptive Cross-Scale Context Module (ACSCM).}

\textbf{Variant - ACSCM1:}  
We adopt a simplified variant, ACSCM1, which omits the latter branch and uses \textit{Recursive Channel-Spatial Attention (RCSA)} instead of CBAM for both current and previous stages. 

We use the version of ACSCM with CBAM Attention for the first four ACSCM blocks and the version of ACSCM with RCSA Attention for the last ACSCM block.

\subsection{Loss Function}

\paragraph{1. Binary Cross-Entropy (BCE) Loss:}
BCE loss is a standard choice for binary segmentation tasks and is defined as:

\begin{align}
\mathcal{L}_{\text{BCE}} = -\frac{1}{N} \sum_{i=1}^N \left[y_i \log(\sigma(p_i)) + (1 - y_i)\log(1 - \sigma(p_i)) \right]
\end{align}

where $p_i$ and $y_i$ are the predicted logit and ground truth for pixel $i$, respectively. $\sigma(\cdot)$ denotes the sigmoid function and $N$ denotes the number of pixels across all images in the batch. BCE ensures correct pixel-wise classification but lacks spatial awareness.

\paragraph{2. IoU Loss:}
Intersection-over-Union (IoU) loss is directly correlated with the performance metric and encourages better region-level prediction:

\begin{align}
\mathcal{L}_{\text{IoU}} = 1 - \frac{\sum_i p_i y_i}{\sum_i p_i + y_i - p_i y_i + \epsilon}
\end{align}

The predicted probability for each pixel is denoted by \( p_i \), where \( p_i \in [0, 1] \), typically obtained by applying a sigmoid activation. The corresponding ground truth label is represented as \( y_i \), where \( y_i \in \{0, 1\} \). The term \( \sum_i p_i y_i \) represents the intersection between the predicted and ground truth masks, summing over the element-wise product of predictions and labels. The union is given by \( \sum_i p_i + y_i - p_i y_i \), accounting for all pixels where either the prediction or the ground truth is positive. To ensure numerical stability and prevent division by zero, a small constant \( \epsilon \) is added to the denominator. The IOU loss complements BCE by enforcing alignment between predicted and ground truth regions, especially useful in cases of unbalanced foreground-background distributions.

\paragraph{3. Focal Tversky (FT) Loss:} 

To further address foreground-background imbalance and emphasize hard-to-classify pixels (e.g., thin or ambiguous regions), we employ the Focal Tversky Loss \cite{abraham2019novel}:

\begin{align}
\mathcal{L}_{\text{FT}} = \left(1 - \frac{\text{TP}}{\text{TP} + \alpha \cdot \text{FP} + \beta \cdot \text{FN} + \epsilon} \right)^\gamma
\end{align}

where TP, FP, and FN represent the number of true positives, false positives, and false negatives, respectively, computed from the predicted probability map \( P \in [0,1]^{H \times W} \) and the ground truth binary mask \( G \in \{0,1\}^{H \times W} \). Specifically:

\begin{align}
\text{TP} &= \sum_{i,j} P_{i,j} \cdot G_{i,j} \\
\text{FP} &= \sum_{i,j} P_{i,j} \cdot (1 - G_{i,j}) \\
\text{FN} &= \sum_{i,j} (1 - P_{i,j}) \cdot G_{i,j}
\end{align}

Here, \( P_{i,j} \) is the predicted foreground probability for pixel \( (i,j) \), and \( G_{i,j} \) is the corresponding ground truth label. 

We set the weighting parameters to \( \alpha = 0.7 \), \( \beta = 0.3 \), and the focusing parameter to \( \gamma = 0.75 \), which biases the loss toward improving recall (by penalizing false negatives more heavily) while also reducing overconfidence in false positives. A small constant \( \epsilon \) is added to the denominator to ensure numerical stability and avoid division by zero.

\subsection{Dataset}

The following datasets were used to test and compare the performance of our method:

\begin{enumerate} 

\item \textbf{DUTS:} 

The DUTS dataset \cite{wang2017learning} contains pixel-level saliency masks representing prominent objects in natural scenes. We use the 10,553 images (DUTS-TR) for training and 5,019 images (DUTS-TE) for testing.

\item \textbf{ECSSD:} 

The Extended Complex Scene Saliency Dataset (ECSSD) \cite{yan2013hierarchical} includes 1,000 images with semantically meaningful objects embedded in cluttered backgrounds. We use this dataset exclusively for testing.

\item \textbf{HKU-IS:} 

HKU-IS \cite{li2015visual} contains 4,447 images with multiple salient objects per image, often exhibiting low contrast with the background. This dataset is used only for testing.

\item \textbf{SBU-Shadow:} 

The SBU Shadow Dataset \cite{vicente2016large} contains natural images with binary shadow masks. We use 4085 images for training and 638 images for testing.

\item \textbf{ISIC2018 (Medical SOD):} 

The ISIC 2018 dataset \cite{codella2019skin} provides dermoscopic skin images with lesion-level annotations. 1886 images are used for training, and 808 images are used for testing. 

\end{enumerate} 

\subsection{Evaluation Metrics}

\paragraph{Mean Absolute Error (MAE).}
MAE computes the average per-pixel absolute difference between the predicted saliency map $\hat{Y}$ and the ground truth mask $Y$:

\begin{equation}
\text{MAE} = \frac{1}{H \times W} \sum_{i=1}^{H} \sum_{j=1}^{W} \left| \hat{Y}_{i,j} - Y_{i,j} \right|,
\end{equation}

where $H$ and $W$ denote the height and width of the image, respectively. A lower MAE indicates higher prediction accuracy.

\paragraph{F-measure $\mathbf{F_{max}}$, $\mathbf{F_{adaptive}}$, and $\mathbf{F_{weighted}}$}
The F-measure combines precision and recall to evaluate the binary classification quality of the saliency map:

\begin{equation}
F_\beta = \frac{(1 + \beta^2) \cdot \text{Precision} \cdot \text{Recall}}{\beta^2 \cdot \text{Precision} + \text{Recall}},
\end{equation}

where $\beta^2$ is commonly set to 0.3 to weigh precision more heavily. We report:

\begin{itemize}
    \item $\mathbf{F_{max}}$: the maximum F-measure over all thresholds.
    \item $\mathbf{F_{adaptive}}$: F-measure computed at an adaptive threshold equal to twice the mean saliency value.
    \item $\mathbf{F_{weighted}}$: a weighted version that considers the imbalance between foreground and background pixels.
\end{itemize}

\paragraph{Structure Measure ($S_\alpha$).}
The Structure Measure evaluates structural similarity between prediction and ground truth by combining object-aware and region-aware structural comparisons:

\begin{equation}
S_\alpha = \alpha \cdot S_o + (1 - \alpha) \cdot S_r,
\end{equation}

where $S_o$ and $S_r$ are object and region similarities, and $\alpha$ balances their contributions ($\alpha = 0.5$).

\paragraph{Enhanced-alignment Measure ($E_\phi$).}
The Enhanced-alignment Measure evaluates the alignment of binary or continuous saliency maps with the ground truth:

\begin{equation}
E_\phi = \frac{1}{W \times H} \sum_{i=1}^{W} \sum_{j=1}^{H} \phi(\hat{Y}_{i,j}, Y_{i,j}),
\end{equation}

where $\phi$ is a local alignment function capturing both global statistics and local pixel consistency.

\paragraph{Intersection over Union (IoU).}
IoU measures the overlap between the predicted binary mask $P$ and the ground truth mask $Y$:

\begin{equation}
\text{IoU} = \frac{|P \cap Y|}{|P \cup Y|}.
\end{equation}

A higher IoU indicates better segmentation overlap.

\paragraph{Computational Efficiency Metrics.}
In addition to accuracy and calibration metrics, it is essential to evaluate the computational complexity and inference efficiency of salient object detection models. These metrics determine the model's suitability for real-time applications, deployment on edge devices, and large-scale processing.

\begin{itemize}
    \item \textbf{Number of Parameters.} Denoted as \texttt{\#Params}, this metric quantifies the total number of learnable weights in the model.

    \item \textbf{Floating Point Operations (FLOPs).} FLOPs measure the total number of multiply-add operations required to process a single image.

    \item \textbf{Inference Time.} Measured in milliseconds (ms), this metric reflects the average time required to forward-pass a single image through the model. 

    \item \textbf{Frames Per Second (FPS).} FPS is the reciprocal of inference time and indicates the real-time processing capability of the model.
\end{itemize}

\end{document}